\documentclass{article}
\usepackage{spconf,amsmath,graphicx}
\ninept 
\usepackage[utf8]{inputenc}
\usepackage{psfrag,epsfig,graphics}
\usepackage{amsmath,amsthm,amssymb,multirow}
\usepackage{mathbbol}
\usepackage{amssymb}             

\DeclareSymbolFontAlphabet{\amsmathbb}{AMSb}%

\usepackage{booktabs}
\usepackage{graphicx}

\usepackage[noadjust]{cite}
\usepackage{multirow}
\usepackage[lined,linesnumbered,ruled]{algorithm2e}

\usepackage{color}  

\newcommand{\cb}[1]{{\boldsymbol{#1}}}
\newcommand{\cp}[1]{\ifmmode {\mathcal{#1}}\else ${\mathcal{#1}}$\fi}

\newcommand{\bA}{\boldsymbol{A}}

\newcommand{\bH}{\boldsymbol{H}}
\newcommand{\bI}{\boldsymbol{I}}

\newcommand{\bM}{\boldsymbol{M}}

\newcommand{\bR}{\boldsymbol{R}}

\newcommand{\bX}{\boldsymbol{X}}

\newcommand{\bm}{\boldsymbol{m}}

\newcommand{\bh}{\boldsymbol{h}}

\newcommand{\be}{\boldsymbol{e}}

\newcommand{\br}{\boldsymbol{r}}

\newcommand{\bx}{\boldsymbol{x}}

\newcommand{\bPsi}{\boldsymbol{\Psi}}

\newcommand{\balpha}{\boldsymbol{\alpha}}

\newcommand{\bbM}{\amsmathbb{M}}
\newcommand{\bbX}{\amsmathbb{X}}
\newcommand{\bbR}{\amsmathbb{R}}
\newcommand{\bbPsi}{\mathbb{\Psi}}

\title{Generalized linear mixing model accounting for endmember variability}
\name{Tales Imbiriba, Ricardo~Augusto~Borsoi, Jos\'e~Carlos~Moreira~Bermudez
\thanks{This work has been supported by the National Council for Scientific and Technological Development (CNPq).}
\thanks{T. Imbiriba, R. A. Borsoi and J.C.M. Bermudez are with the Department of Electrical Engineering, Federal University of Santa Catarina, Florian\'opolis, SC, Brazil. e-mail: talesim@gmail.com; raborsoi@ucs.br; j.bermudez@ieee.org.}%
}
\address{Federal University of Santa Catarina, Florianópolis, SC, Brazil}
\begin{document}
%
\maketitle
\begin{abstract}
Endmember variability is an important factor for accurately unveiling vital information relating the pure materials and their distribution in hyperspectral images. Recently, the extended linear mixing model (ELMM) has been proposed as a modification of the linear mixing model (LMM) to consider endmember variability effects resulting mainly from illumination changes. In this paper, we further generalize the ELMM leading to a new model (GLMM) to account for more complex spectral distortions where different wavelength intervals can be affected unevenly. 
We also extend the existing methodology to jointly estimate the variability and the abundances for the GLMM. Simulations with real and synthetic data show that the unmixing process can benefit from the extra flexibility introduced by the GLMM.
\end{abstract}
\section{Introduction}
\label{sec:intro}

Hyperspectral imaging has attracted formidable interest of the scientific community in the past two decades, where hyperspectral images (HIs) have been explored in an increasing number of applications in different fields~\cite{Bioucas-Dias-2013-ID307}. The limited spatial resolution of hyperspectral devices often mixes the spectral contribution of different pure materials, termed \emph{endmembers}, in the scene~\cite{Keshava:2002p5667}. This phenomenon is more prominent in remote sense applications due to the distance between airborne and spaceborne sensors and the target scene. The mixing process must be well understood to accurately unveil vital information relating the presence of pure materials and their distribution in the scene. 
Hyperspectral unmixing (HU) aims to solve this problem by decomposing the hyperspectral image (HI) into a collection of endmembers and their fractional abundances~\cite{Bioucas2012}. 

Different mixing models have been proposed to explain the interaction between light and the endmembers~\cite{Dobigeon-2014-ID322}.
The simplest and most widely used model is the Linear Mixing Model (LMM)~\cite{Keshava:2002p5667}. The LMM assumes that the observed reflectance vector (\emph{i.e.} a HI pixel) can be modeled as a convex combination of the spectral signatures of the endmembers present in the scene. This assumption imposes positivity and sum-to-one constraints on the linear combination coefficients, leading to their interpretation as fractional abundances of the corresponding endmembers in each pixel. The simplicity of the LMM and the convexity constraints on the fractional abundances naturally lead to fast and reliable unmixing strategies. Given the intrinsic limitations of this simple mixing model, more refined unmixing approaches have been more recently proposed to account nonideal effects such as nonlinearity~\cite{Dobigeon-2014-ID322,Imbiriba2016_tip, Imbiriba2017_bs_tip} and endmember variability, often found in a typical scenes~\cite{Zare-2014-ID324, drumetz2016variabilityReviewRecent}.

Endmember variability can be caused by a myriad of factors including environmental, illumination, atmospheric and temporal changes~\cite{Zare-2014-ID324}, and its occurrence may result in significant estimation errors being propagated
throughout the unmixing process~\cite{Thouvenin_IEEE_TSP_2016}. The most common approaches to deal with spectral variability can be divided in three basic classes. 1) to group endmembers in variational sets, 2) to model endmembers as statistical distributions, and 3) to incorporate the variability in the mixing model, often using physically motivated concepts~\cite{drumetz2016variabilityReviewRecent}. This work follows the third approach. Recently, \cite{Thouvenin_IEEE_TSP_2016} and \cite{drumetz2016blind} introduced variations of the LMM to cope with  spectral variability. 
The model called Perturbed LMM model (PLMM) introduced in~\cite{Thouvenin_IEEE_TSP_2016} defined an additive perturbation to the endmember matrix. Such perturbation matrix then needs to be estimated jointly with the abundances. Though the perturbation matrix can model arbitrary endmember variations, it lacks physical motivation. 
The Extended Linear Mixing Model (ELMM) proposed in~\cite{drumetz2016blind} extended the LMM model by using one pixel-dependent multiplicative term for each endmember, a generalization that can efficiently model changes in the observed reflectances due to illumination, an important effect~\cite{drumetz2016blind}. 
This model addresses a physically motivated problem, with the advantage of estimating a variability parameter vector of much lower dimension when compared with the additive perturbation matrix in PLMM.
Although the ELMM performs well in situations where spectral variability is mainly caused by illumination variations, it lacks a necessary flexibility when the endmembers are subject to more complex spectral distortions. For instance, experimental measurements on vegetation spectra under different conditions have shown a significant dependence of the spectral variation on wavelength intervals~\cite{gao2006multiSeasonalSaltMarshChina,schmidt2000variabilityVegetationArid,lukevs2013variabilityLeaves}. This type of variability is not supported by the ELMM model in~\cite{drumetz2016blind}, which assumes a fixed scaling across all wavelengths.

In this work we introduce a generalization of the ELMM model proposed in~\cite{drumetz2016blind} to accounts for endmember variability in arbitrary regions of the measured spectrum. We call the resulting model the Generalized Linear Mixing Model (GLMM). 
The estimation of the required parameters is realized by generalizing the methodology used in~\cite{drumetz2016blind} through the use of three-dimensional tensors to accommodate the new model without significantly effecting the simplicity of the proposed solution or its computational complexity.
Simulation results using synthetic and real data indicate that the extra flexibility introduced by the GLMM model can improve the results of existing methods for different types of endmember variability. 

This paper is organized as follows. In Section~\ref{sec:LMMs} we briefly review the LMM and its extended version ELMM. Section~\ref{sec:GLMM} introduces the proposed GLMM. In Section~\ref{sec:Problem} we define new tensor variables and extend the solution in~\cite{drumetz2016blind} to the GLMM. The performance of the proposed method is compared with competing algorithms in Section~\ref{sec:Simulations}. Finally, the conclusions are presented in Section~\ref{sec:conclusions}.

\section{Extended Linear Mixing Model}\label{sec:LMMs}

The Linear Mixing Model (LMM)~\cite{Keshava:2002p5667} assumes that a given a pixel $\br_n = [r_{n,1},\,\ldots, \,r_{n,L} ]^\top$, with $L$ bands, is represented as
\begin{equation}
\begin{split}
 &\br_n = \bM \balpha_n + \be_n\\
 &\text{subject to }\,\cb{1}^\top\balpha_n = 1 \text{ and } \balpha_n \succeq \cb{0}
\end{split}
\label{eq:LMM}
\end{equation}
where $\bM = [\bm_1,\,\ldots, \,\bm_R]$ is an $L\times R$ matrix whose columns are the $R$ endmember spectral signatures $\bm_i = [m_{i,1},\,\ldots,\,m_{i,L}]^\top$, $\balpha_n = [\alpha_{n,1},\,\ldots,\,\alpha_{n,R}]^\top$ is the abundance vector, $\be\sim\mathcal{N}(0, \sigma_n^2\bI_L)$ is an additive white Gaussian noise (WGN), $\bI_L$ is the $L\times L$ identity matrix, and $\succeq$ is the entrywise $\geq$ operator. 
The LMM assumes that the endmember spectra are fixed for all pixels $\br_n$, $n=1,\ldots,N$, in the HI. This assumption can jeopardize the accuracy of estimated abundances in many circumstances due to the spectral variability existing in a typical scene.
The Extended Linear Mixing Model (ELMM)~\cite{drumetz2016blind} partially mitigates such limitation by including a multiplicative diagonal weight matrix  $\cb{\psi}_n = \text{diag}(\psi_{1,n},\ldots,\psi_{R,n})$ in the LMM such that
\begin{equation} \label{eq:elmm_model_i}
\br_n = \bM \cb{\psi}_n \balpha_n + \be_n
\end{equation}
with $\psi_{k,n}\in\amsmathbb{R}^{+},\,k=1,\ldots,R$. Each coefficient $\psi_{k,n}$ scales the whole spectrum of endmember $\bm_i$ in pixel $n$, leading to a simple strategy to model variability resulting from illumination effects. 


\section{Generalized Linear Mixing Model (GLMM)}~\label{sec:GLMM}
As explained in Section~1, we propose a generalization of the ELMM model to allow for spectral variabilities per wavelength intervals. To this end, we propose to employ a band-dependent scaling factor, enabling the new model to adapt to arbitrary variations of the endmember spectra. In the new GLMM model each pixel $\br_n$ is written as
\begin{equation}
\br_n = (\bM\odot\cb{\Psi}_n)\balpha_n + \be_n
\end{equation}
where $\cb{\Psi}_n\in{\amsmathbb{R}^{L\times R}}$ is a scaling matrix with entries $[\cb{\Psi}_n]_{\ell,k} = \psi_{n_{\ell,k}}\geq 0$, and $\odot$ is the Hadamard product.
This model is a generalization of the ELMM where the scaling matrix $\bPsi_n$ acts on each wavelength of each endmember individually. Such feature leads to a more flexible model that allows to consider variabilities that are not uniform along each endmember spectrum. ELMM is clearly a particular case of GLMM, and the new model can be employed for any level of granularity of variability per wavelength ranges, to the limit of an independent scaling of each wavelength component of each endmember in each pixel.   

\section{The Unimixing Problem}\label{sec:Problem}
Assuming the availability of a reference endmember matrix $\bM_0$ (which can be obtained using any endmember extraction method), the HU problem reduces to estimating the free parameters minimizing a given risk functional defined for the whole HI $\bR = [\br_1,\ldots,\br_N]$. For this purpose the methodology  presented in~\cite{drumetz2016blind} can be extended for the GLMM by defining three-dimensional tensors. Thus, we propose to minimize the following regularized cost functional:
\begin{equation}
\begin{split}
 J(\bA,\cb{\bbM},\cb{\bbPsi}) =& \frac{1}{2}\sum_{n=1}^{N}\left(\|\br_n-\bM_n\balpha_n\|^2 \right.\\
 &\left.+ \lambda_M\|\bM_n-\bM_0\odot\bPsi_n\|^2_{F} \right) \\
 &+ \mathcal{R}(\bA)+\mathcal{R}(\bbPsi).
\end{split}
\label{eq:glmm_cost_func}
\end{equation}
where $\bbM$ and $\bbPsi$ are $L\times R\times N$ tensors with entries $[\bbM]_{:,:,n} = \bM_n$, and $[\bbPsi]_{:,:,n}=\bPsi_n$ respectively, $\bA=[\balpha_1,\ldots,\balpha_N]$ is the abundance matrix, $\lambda_M$ is a parameter that controls the strictness of the regularization over $\bM_n$, and $\mathcal{R}(\bA)$ and $\mathcal{R}(\bbPsi)$ are spatial regularizations over $\bA$ and $\bbPsi$.
Thus, the optimization problem becomes
\begin{equation}
(\bA^*,\bbM^*,\bbPsi^*) = \mathop{\arg\min}_{\bA,\,\cb{\bbM},\,\cb{\bbPsi}} J(\bA,\cb{\bbM},\cb{\bbPsi}).
\label{eq:opt_glmm}
\end{equation}
The problem defined in~\eqref{eq:opt_glmm} is non-smooth and non-convex with respect to all variables $\bA$, $\bbM$, and $\bbPsi$, but is convex with respect to each one of them. Thus, we follow the same approach used in~\cite{drumetz2016blind} and find a local stationary point minimizing~\eqref{eq:opt_glmm} iteratively with respect to each variable, leading to the strategy presented in Algorithm~\ref{alg:global_opt}.

\begin{algorithm} [bth]
\SetKwInOut{Input}{Input}
\SetKwInOut{Output}{Output}
\caption{Global algorithm for solving \eqref{eq:glmm_cost_func}~\label{alg:global_opt}}
\Input{$\bR$, $\lambda_M$, $\lambda_A$, $\lambda_{\bbPsi}$, $\bA^{(0)}$, $\bbPsi^{(0)}$ and $\bM_0$.}
\Output{$\bA^*$, $\bbM^*$ and $\bbPsi^*$.}
Set $i=0$ \;
\While{stopping criterion is not satisfied}{
$i=i+1$ \;
$\bbM^{(i)} = \underset{\bbM}{\arg\min} \,\,\,\,  {J}(\bA^{(i-1)},\bbM,\bbPsi^{(i-1)})$ \;
$\bA^{(i)} = \underset{\bA}{\arg\min} \,\,\,\,  {J}(\bA,\bbM^{(i)},\bbPsi^{(i-1)})$ \;
$\bbPsi^{(i)} = \underset{\bbPsi}{\arg\min} \,\,\,\,  {J}(\bA^{(i)},\bbM^{(i)},\bbPsi)$ \;
}
\KwRet $\bA^*=\bA^{(i)}$,~ $\bbM^*=\bbM^{(i)}$,~ $\bbPsi^*=\bbPsi^{(i)}$  \;
\end{algorithm}

The regularization functionals $\mathcal{R}(\bA)$ and $\mathcal{R}(\bbPsi)$ in~\eqref{eq:glmm_cost_func} are selected in order to provide spatial smoothness to the abundances and scaling factors and to enforce physical constraints (i.e. positivity and sum-to-one constraints on the abundances). They are selected as
\begin{align}
	\mathcal{R}(\bA) {}={} & \lambda_A \big(\| \mathcal{H}_h(\bA)\|_{2,1} + \| \mathcal{H}_v(\bA)\|_{2,1} \big) \notag\\
    &+ \iota_{\bbR_+}(\bA) + \cb{\mu}^\top (\bA^\top \cb{1}_{R\times 1} - \cb{1}_{N\times 1})\notag\ 
\end{align}
and
\begin{align}
	\mathcal{R}(\bbPsi) {}={} & 
    \frac{\lambda_\bbPsi}{2} \sum_{\ell=1}^L \sum_{k=1}^R (\|\cp{H}_h([\bbPsi]_{\ell,k,:})\|^2 
    + \|\cp{H}_v([\bbPsi]_{\ell,k,:})\|^2) \notag
\end{align}
where $[\cdot]_{\ell,k,:}$ is a slice of a tensor for band $\ell$, endmember $k$ and all $N$ pixels, and the parameters $\lambda_A$ and $\lambda_\bbPsi$ control the weights of the regularization terms in the cost function.
The linear operators $\mathcal{H}_h$ and $\mathcal{H}_v$ compute the first-order horizontal and vertical gradients of a bidimensional signal, acting separately for each material of $\bA$.
The spatial regularization in the abundances is promoted by a mixed $\mathcal{L}_{2,1}$ norm of their gradient, where $\|\bX\|_{2,1}=\sum_{n=1}^N\|\bx_n\|_2$. This norm is used to promote sparsity of the gradient across different materials (i.e. to force neighboring pixels to be homogeneous in all constituent endmembers). The $\mathcal{L}_1$ norm can also be used, leading to the Total Variation regularization, where $\|\bX\|_{1}=\sum_{n=1}^N\|\bx_n\|_1$~\cite{iordache2012total}.
$\iota_{\bbR_+}(\cdot)$ is the indicator function of $\bbR_+$ (i.e. $\iota_{\bbR_+}(a)=0$ if $a\geq0$ and $\iota_{\bbR_+}(a)=\infty$ if $a<0$) acting component-wise on its input, and enforces the abundances positivity constraint.
$\cb{\mu}\in\bbR^N$ is a vector of Lagrange multipliers associated with the sum-to-one constraint. Note that the cost function must also be optimized with respect to $\cb{\mu}$ for this constraint to be enforced.

\subsection{Optimiztion with respect to $\bbM$}
Rewriting the problem~\eqref{eq:opt_glmm} using only the terms in~\eqref{eq:glmm_cost_func} that depend on $\bbM$, the problem becomes
\begin{equation}
\begin{split}
\bbM^* =& \mathop{\arg \min}_{\bbM \succeq \cb{0}} \frac{1}{2}\sum_{n=1}^N \left(\|\br_n-\bM_n\balpha_n\|^2\right.\\ 
&\left. - \lambda_M\|\bM_n -\bM_0\odot\bPsi_n\|^2_F\right).
\end{split}
\label{eq:M_problem}
\end{equation}
The problem in~\eqref{eq:M_problem} can be solved individually for each pixel $\br_n$. Thus, relaxing the positivity constraint on the elements of $\bbM$, the solution can be found as
\begin{equation}
\bM_n^* = (\br_n\balpha^\top+ \lambda_M\bM_0\odot\bPsi_n)(\balpha_n\balpha_n^\top + \lambda_s\bI_R)^{-1}
\end{equation}
where $\bI_R$ is the $R\times R$ identity matrix. Then, an approximate solution to the constrained problem can be obtained by projecting $\bM_n^*$ onto the nonnegative orthant $\bbR_+^{L\times R}$ by thresholding the negative entries to zero~\cite{drumetz2016blind}.

\subsection{Optimization with respect to $\bA$}
Restating the problem~\eqref{eq:opt_glmm} only considering the terms in~\eqref{eq:glmm_cost_func} that depend on $\bA$ leads to the following optimization problem for the abundance matrix
\begin{equation}
\begin{split}
 \bA^* {}={} &  \mathop{\arg\min}_{\bA} \frac{1}{2} \sum_{n=1}^{N} \|\br_n - \bM_n\balpha_n\|^2\\
 &+ \lambda_A ( \|\cp{H}_h(\bA)\|^2_{2,1} + \|\cp{H}_v(\bA)\|^2_{2,1}) \\
 &+ \iota_{\bbR_+}(\bA) + \cb{\mu}^\top (\bA^\top \cb{1}_{R\times 1} - \cb{1}_{N\times 1})
\end{split}
\label{eq:A_problem}
\end{equation}
This problem is clearly not separable with respect to the pixels in the image. However, problem~\eqref{eq:A_problem} can be efficiently solved using the Alternating Direction Method of the Multipliers (ADMM)~\cite{Boyd_admm_2011}. The procedure is well described in~\cite{drumetz2016blind} and will be suppressed here for conciseness.

\subsection{Optimization with respect to $\bbPsi$}
Rewriting the optimization problem~\eqref{eq:opt_glmm} considering only the terms in~\eqref{eq:glmm_cost_func} that depend on $\bbPsi$ leads to
\begin{equation}
\begin{split}
 \bbPsi^* =& \mathop{\arg\min}_{\bbPsi} \frac{\lambda_M}{2}\sum_{n=1}^N\|\bM_n-\bM_0\odot\bPsi_n\|^2_F\\ 
 &+ \frac{\lambda_\bbPsi}{2}(\|\cp{H}_h(\bbPsi)\|^2_F + \|\cp{H}_v(\bbPsi)\|^2_F)
\end{split}
\end{equation}
which can be rewritten as 
\begin{equation}
\begin{split}
\bbPsi^* =& \mathop{\arg\min}_{\bbPsi} \frac{\lambda_M}{2}\sum_{\ell=1}^{L}\sum_{k=1}^R\left( \|[\bbM]_{\ell,k,:} - m_{\ell,k}^0[\bbPsi]_{\ell,k,:}\|^2\right) \\
 &+ \frac{\lambda_\bbPsi}{2}(\|\cp{H}_h([\bbPsi]_{\ell,k,:})\|^2 + \|\cp{H}_v([\bbPsi]_{\ell,k,:})\|^2)
\end{split}
\end{equation}
where $[\cdot]_{\ell,k,:}$ is a slice of a tensor for band $\ell$, endmember $k$ and all $N$ pixels and $m_{\ell,k}^0 = [\bM_0]_{\ell,k}$ is a scalar. 
The problem can be solved for each endmember $k$ and band $\ell$ individually, and its solution is given by
\begin{equation}
[\bbPsi]_{\ell,k,:}^* = \cb{P}^{-1}\left(\lambda_M m_{\ell,k}^0 [\bbM]_{\ell,k,:}\right)
\label{eq:sol_Psi}
\end{equation}
where $\cb{P} = \lambda_M (m_{\ell,k}^0)^2\bI_N + \lambda_{\bbPsi}(\bH_h^\top\bH_h + \bH_v^\top\bH_v)$.
The solution in~\eqref{eq:sol_Psi} involves the inverse of the $N\times N$ matrix $\cb{P}$ which can be computationally intensive or intractable. However, if we assume periodic boundary conditions for the differential operators $\cp{H}_h$ and $\cp{H}_v$, the corresponding matrices $\bH_v$, $\bH_h$ and consequently $\cb{P}$ will have the structure of a block circulant matrix with circulant blocks (BCCB). Since BCCB matrices can be diagonalized using the bi-dimensional Discrete Fourier Transform, problem~\eqref{eq:sol_Psi} can be solved efficiently as follows~\cite{hansen2006deblurringBook} 
\begin{equation}
[\bbPsi]_{\ell,k,:}^* = \cp{F}^{-1}\left( \frac{\cp{F}(\lambda_M m_{\ell,k}^0 [\bbM]_{\ell,k,:})}{\lambda_M (m_{\ell,k}^0)^2\cb{1}_{p\times q} + \lambda_{\bbPsi}(|\cp{F}(\bh_h)|^2 + |\cp{F}(\bh_v)|^2 )}  \right)\nonumber
\end{equation}
where $\cp{F}$ and $\cp{F}^{-1}$ represents the bi-dimensional discrete Fourier transform and its inverse respectively, $\cb{1}_{p\times q}$ is a $p\times q$ matrix of ones, where $p$ and $q$ are the number of rows and columns of the HI cube, and $\bh_h$ and $\bh_v$ are convolution masks represented as $p\times q$ matrices.

\begin{table*}[htb!]
\footnotesize
\caption{Simulations with synthetic data.}
\begin{center}
\renewcommand{\arraystretch}{1}
\begin{tabular}{lccccc|ccccc|cc}
\toprule
\multicolumn{6}{c}{Data Cube 0 -- DC0} & \multicolumn{5}{|c}{Data Cube 1 -- DC1} &\multicolumn{2}{|c}{Houston Data}\\
\toprule\bottomrule
& $\text{RMSE}_{\bA}$ & $\text{RMSE}_{\bbM}$ &$\text{SAM}_{\bbM}$ & $\text{RMSE}_{\bR}$ &$\text{SAM}_{\bR}$ & $\text{RMSE}_{\bA}$ & $\text{RMSE}_{\bbM}$ &$\text{SAM}_{\bbM}$ & $\text{RMSE}_{\bR}$ &$\text{SAM}_{\bR}$ & $\text{RMSE}_{\bR}$ &$\text{SAM}_{\bR}$\\ \midrule
FCLS & 0.0968 & - & - & 0.0420 & 0.0566 & 0.0243 & - & - &0.0385 & 0.0492 & 0.006082 & 0.057695 \\
SCLS & 0.0642 & 0.0673 & 0.0625 & 0.0403 & 0.0548 & 0.0509 & 0.0457 & 0.0617 & 0.0383 & 0.0489 &  0.006082 & 0.025361\\
PLMM & 0.0641 & 0.0689 & \textbf{0.0566} & 0.0250 & 0.0341 & 0.0476 & 0.0401 & 0.0578 & 0.0257 & 0.0329 & 0.002918 & 0.013617\\
ELMM & 0.0540 & 0.0608 & 0.0568 & 0.0254 & 0.0346 & 0.0209 & 0.0425 & 0.0609 & 0.0367 & 0.0469 & 0.003217 & 0.013225\\
GLMM & \textbf{0.0512} & \textbf{0.0587} & 0.0601 & \textbf{0.0011} & \textbf{0.0011} & \textbf{0.0193} & \textbf{0.0391} & \textbf{0.0564} & \textbf{0.0202} & \textbf{0.0257} & \textbf{0.000313} & \textbf{0.000472}\\
\bottomrule\toprule
\end{tabular}
\end{center}
\label{tab:results_synthData}
\end{table*}

\section{Simulations}\label{sec:Simulations}
In this section, the performance of the proposed methodology is illustrated through simulations with both synthetic and real data. We compare the proposed method based on the GLMM with the the fully constrained least squares (FCLS), the scaled constrained least squares (SCLS), the PLMM~\cite{Thouvenin_IEEE_TSP_2016}, and the ELMM~\cite{drumetz2016blind}.



To measure the accuracy of the unmixing methods we consider the Root Means Squared Error (RMSE)
\begin{equation}
\text{RMSE}_{\bbX} = \sqrt{\frac{1}{N_{\bbX}}\|\text{vec}(\bbX)- \text{vec}(\left.\bbX\right.^*)\|^2} 
\end{equation}
where $\text{vec}(\cdot)$ is the vectorization operator $\bbX\rightarrow \bx$, $\amsmathbb{R}^{a\times b\times c}\mapsto \amsmathbb{R}^{abc}$, $N_{\bbX} = abc$. In this work we apply the RMSE to evaluate the estimates of the abundances ($\text{RMSE}_{\bA}$), of the endmembers tensor ($\text{RMSE}_{\bbM}$) and of the reconstructed images ($\text{RMSE}_{\bR}$).
We also consider the Spectral Angle Mapper for the HI



\begin{equation}
  \text{SAM}_{\bR} = \frac{1}{N}\sum_{n=1}^{N}\arccos\left(\frac{\br_n^\top\br_n^*}{\|\br_n\|\|\br^*_n\|}\right)
\end{equation}
and for the endmembers tensor
\begin{equation}
  \text{SAM}_{\bbM} = \frac{1}{N}\sum_{n=1}^{N}\sum_{k=1}^{R}\arccos\left(\frac{\bm_{k,n}^\top\bm_{k,n}^*}{\|\bm_{k,n}\|\|\bm_{k,n}^*\|}\right).
\end{equation}

\begin{figure}[htb]
\centering
\includegraphics[width=0.15\textwidth]{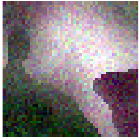}\qquad\qquad
\includegraphics[width=0.15\textwidth]{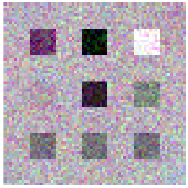}
\caption{Synthetic data cubes DC0, left, and DC1, right. }\label{fig:synthDatas}
\end{figure}

\subsection{Synthetic data}
 For a comprehensive comparison among the different methods we created two synthetic datasets, namely Data Cube 0 (DC0) and Data Cube 1 (DC1), represented in Figure~\ref{fig:synthDatas}. These datasets were built using endmembers extracted from the USGS Spectral Library~\cite{clark2003imaging}, and different strategies were used to generate the abundance maps, which exhibit spatial correlation between neighboring pixels. For DC0, we adopted the variability model used in~\cite{drumetz2016blind} (a multiplicative factor acting in each endmember), while for DC1 we considered the variability following the GLMM where correlation was imposed over $\bPsi_n$ using a 3-D Gaussian filter. White Gaussian noise was added to both datasets resulting in a SNR of 30dB.

To find the optimal parameters for the selected algorithms we performed a grid search for each dataset. The parameter ranges were chosen based on the ranges tested and discussed by the authors in the original publication of each algorithm. For the PLMM we used $\gamma=1$, since the authors fixed this parameter in all simulations, and searched for $\alpha$ and $\beta$ in the range $[0.35,\, 0.7,\, 1.4,\, 25]$ and $[10^{-9},\, 10^{-5},\, 10^{-4},\, 10^{-3}]$, respectively. For both the ELMM and the GLMM, the parameters were selected among the following values: $\lambda_{S},\,\lambda_M \in [0.01,\, 0.1,\, 1,\, 5,\, 10,\, 15]$, $\lambda_{A} \in [0.001,\, 0.01,\, 0.05]$, and $\lambda_\psi,\,\lambda_{\bbPsi} \in [10^{-6},\, 10^{-3},\, 10^{-1}]$.

The results are presented in Table~\ref{tab:results_synthData}. In terms of RMSE for the abundance vectors, $\text{RMSE}_{\bA}$, the proposed strategy clearly outperformed the competing algorithms for both datasets. This behavior can be verified for almost all metrics considered. The only exception is the $\text{SAM}_{\bbM}$ for DC0 where PLMM and ELMM presented smaller spectral angles.
Regarding the increase of computational complexity introduced by the GLMM when compared with the ELMM, the simulations point out that the GLMM approach demanded $3.62\times \text{Time}_{\text{ELMM}}$ for DC0 and $1.92\times \text{Time}_{\text{ELMM}}$ for DC1, where $\text{Time}_{\text{ELMM}}$ is the CPU time elapsed during the ELMM unmixing process.
The results show that the extra flexibility of the GLMM can be beneficial for the HU problem at the expense of a reasonable increase in the computational complexity.

\begin{figure}[htb]
\centering
\includegraphics[height=7.8cm ,width=0.45\textwidth]{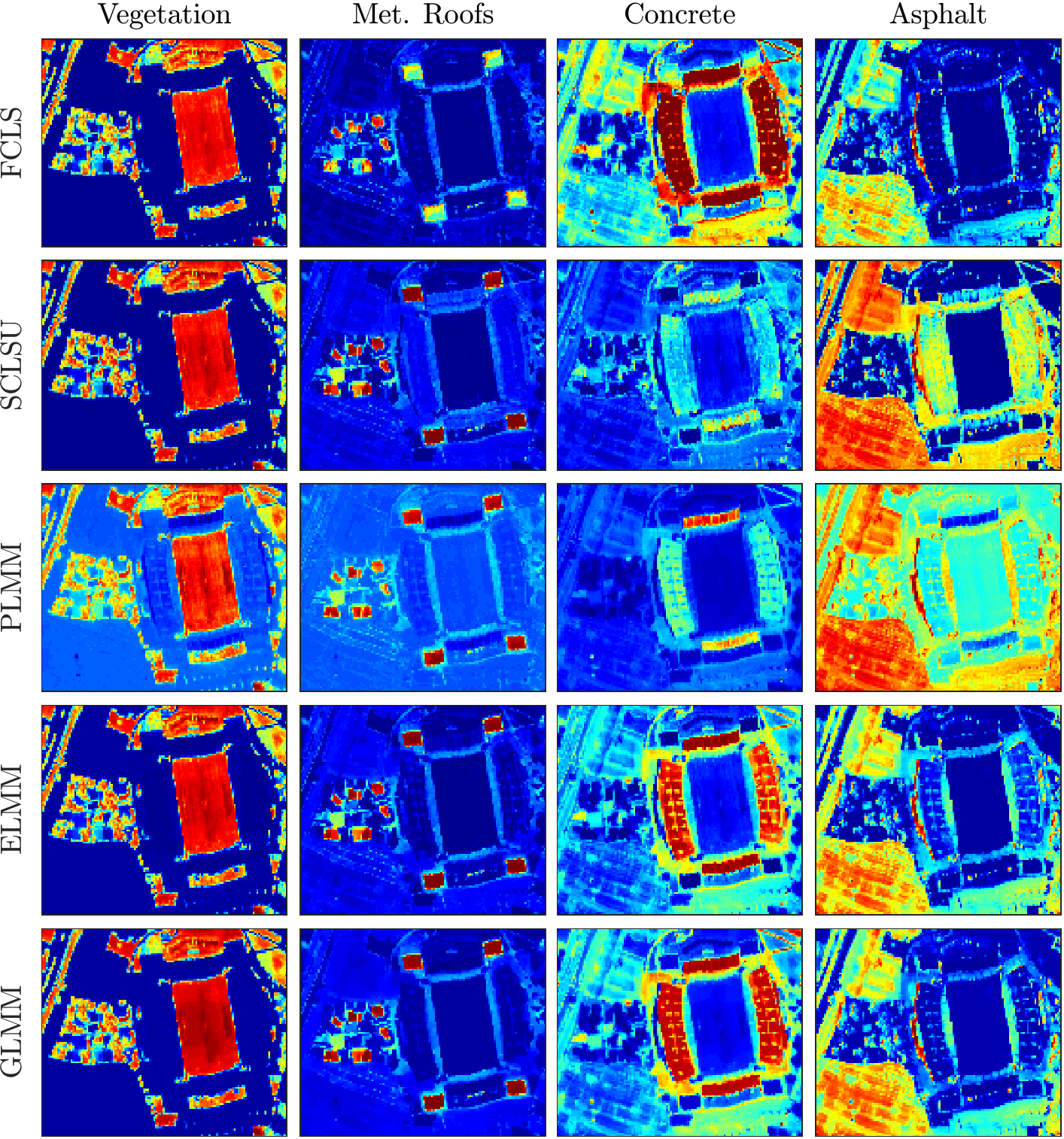}
\caption{Abundance maps of the Houston dataset for all tested algotithms where the abundance values are represented by colors ranging from blue ($\alpha_k = 0$) to red ($\alpha_k = 1$).}\label{fig:ab_maps_houston}
\end{figure}

\subsection{Real data}
For simulations with real data we considered the Houston dataset discussed in~\cite{drumetz2016blind}. This dataset is known to have four endmembers which were extracted using the VCA algorithm~\cite{Nascimento2005}. Figure~\ref{fig:ab_maps_houston} shows the reconstructed abundance maps for all tested methods while Table~\ref{tab:results_synthData} presents the results in terms of $\text{RMSE}_{\bR}$ and $\text{SAM}_{\bR}$. 
Figure~\ref{fig:ab_maps_houston} shows that the proposed GLMM method provided smooth and accurate abundance estimation, comparable with the results obtained using the ELMM. In fact, for the Concrete endmember, the GLMM abundance map shows stronger components in the stadium stands when compared with the other methods considering spectral variability. Although the results presented in Table~\ref{tab:results_synthData} indicate better fitting for the GLMM method in both RMSE and SAM, these results should be taken with the proper care, since the connection of reconstruction error and abundance estimation is not straightforward. The GLMM demanded a computational time of $0.92\times \text{Time}_{\text{ELMM}}$.



\section{Conclusions}\label{sec:conclusions}

This paper proposed a new Generalized Linear Mixing Model (GLMM) that accounts for endmember spectral variability. The new model generalizes the Extended Linear Mixing Model (ELMM) to allow for the consideration of band dependent scaling factors for the endmember signatures. This way the GLMM model can represent a larger variety of realistic spectral variations of the endmembers, generalizing the representation capability of the ELMM.  To solve the resulting optimization problem, we extended the variable splitting methodology used in~\cite{drumetz2016blind} by including new tensor variables. Simulation results with both synthetic and real data suggest that the extra flexibility introduced by the GLMM can be beneficial for the unmixing process, resulting in improvements in both the abundance estimation and the reconstruction error.


\bibliographystyle{IEEEbib}
\bibliography{strings,hyperspectral}

\end{document}